\documentclass[10pt,twocolumn,letterpaper]{article}

\usepackage[pagenumbers]{wacv}

\usepackage{caption, cuted}
\usepackage{lipsum}
\usepackage{makecell}
\usepackage{graphicx}
\usepackage{algorithm}
\usepackage{algpseudocode}
\usepackage{flushend}
\usepackage{booktabs}
\usepackage{svg}
\usepackage{siunitx}
\usepackage{balance}

\newcommand{\ours}{\textit{SceneEdited}}

\definecolor{wacvblue}{rgb}{0.21,0.49,0.74}
\usepackage[pagebackref,breaklinks,colorlinks,allcolors=wacvblue]{hyperref}

\usepackage{standalone}

\def\wacvPaperID{835} 
\def\confName{WACV}
\def\confYear{2026}

\title{SceneEdited: A City-Scale Benchmark for 3D HD Map Updating via Image-Guided Change Detection}

\author{
Chun-Jung Lin\qquad
Tat-Jun Chin\qquad
Sourav Garg\qquad
Feras Dayoub\\
Australian Institute for Machine Learning (AIML), University of Adelaide, Australia\\
}

\begin{document}
\maketitle

\begin{strip}
\vspace*{-1.5cm}
  \centering
  \includegraphics[width=\textwidth]{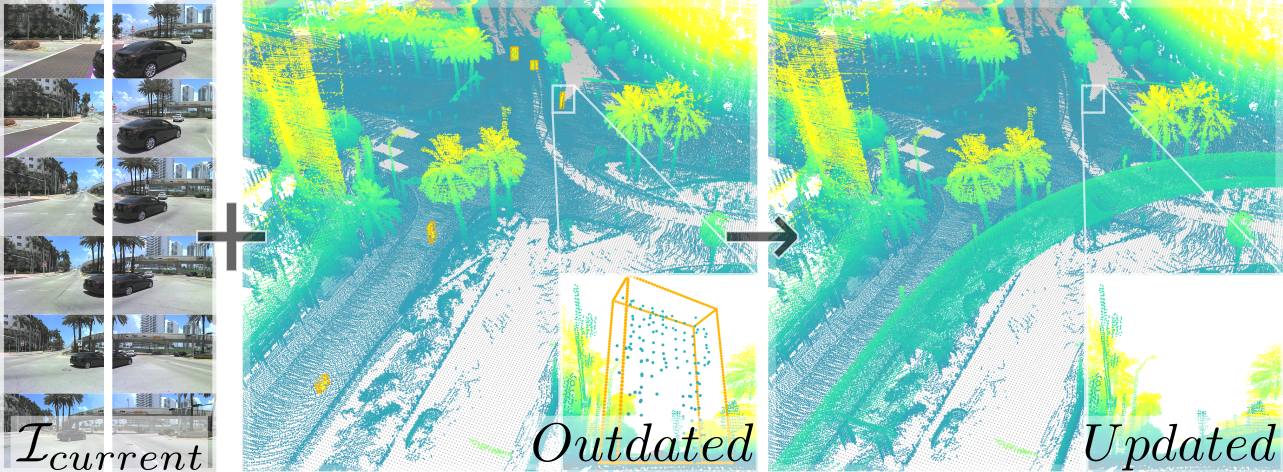}
  \captionof{figure}{\textbf{\ours{} task overview.} RGB images show the current urban scene, whereas the stored 3-D HD map has been synthetically altered by adding and removing objects. Our benchmark challenges methods to detect these changes from the images and automatically update the 3-D map.}
  \label{fig:widefirst}
\end{strip}
\begin{abstract}


Accurate, up-to-date High-Definition (HD) maps are critical for urban planning, infrastructure monitoring, and autonomous navigation. However, these maps quickly become outdated as environments evolve, creating a need for robust methods that not only detect changes but also incorporate them into updated 3D representations. While change detection techniques have advanced significantly, there remains a clear gap between detecting changes and actually updating 3D maps, particularly when relying on 2D image-based change detection. To address this gap, we introduce \textbf{SceneEdited}, the first city-scale dataset explicitly designed to support research on HD map maintenance through 3D point cloud updating. \textbf{SceneEdited} contains over 800 up-to-date scenes covering 73 km of driving and approximate 3~$\text{km}^2$ of urban area, with more than 23,000 synthesized object changes created both manually and automatically across 2000+ out-of-date versions, simulating realistic urban modifications such as missing roadside infrastructure, buildings, overpasses, and utility poles. Each scene includes calibrated RGB images, LiDAR scans, and detailed change masks for training and evaluation. We also provide baseline methods using a foundational image-based structure-from-motion pipeline for updating outdated scenes, as well as a comprehensive toolkit supporting \textbf{scalability}, \textbf{trackability}, and \textbf{portability} for future dataset expansion and unification of out-of-date object annotations. Both the dataset and the toolkit are publicly available at \url{https://github.com/ChadLin9596/ScenePoint-ETK}, establising a standardized benchmark for 3D map updating research.

\end{abstract}

\section{Introduction}
\label{sec:intro}

Maintaining accurate, up-to-date city-scale High-Definition (HD) maps is essential for a wide range of real-world applications, including autonomous driving~\cite{jo2018simultaneous, pannen2020keep}, urban planning~\cite{elghazaly2023high}, infrastructure monitoring~\cite{wijaya2023multi}, and navigation systems~\cite{sun2025mind}. These maps serve as critical priors for situational awareness, motion planning, and safe decision making~\cite{sun2025mind} in dynamic environments. An outdated map that fails to capture new buildings, road modifications, or removed obstacles can significantly degrade system performance~\cite{elghazaly2023high}, leading to safety risks and operational failures~\cite{pannen2019hd}. Ensuring the reliability and currency of 3D maps is therefore not only a matter of efficiency but also of safety, trust, and practical deployment at scale. However, producing HD maps relies on professional and expensive surveying systems and time-consuming manual work, making frequent full-scale rebuilding impractical~\cite{elghazaly2023high, wijaya2024high, wijaya2023multi}. Instead, map maintenance focuses on updating existing maps selectively, enabling cost-effective, timely, and scalable solutions.

HD maps are structured as multi-layer representations tailored to different applications~\cite{elghazaly2023high}, ranging from the point cloud base map layer, which provides dense 3D geospatial data of roads and infrastructure, to the semantic map layer that captures road features such as traffic lights, crosswalks, and lane geometries, and a real-time information layer encoding dynamic elements like construction zones or traffic flow. Such layered architectures are essential and have been widely adopted by commercial mapping companies including HERE~\cite{here2017live}, TomTom, as well as by self-driving companies such as Waymo~\cite{Sun_2020_CVPR}, Argo AI~\cite{chang2019argoverse}, Motional~\cite{nuscenes}, and Lyft~\cite{houston2021one}. However, this layered design also introduces distinct challenges for HD map maintenance, as each layer requires tailored strategies for detecting and updating changes. While the semantic layer has been the focus of substantial research and development in change detection and updating methods, support for updating the Point Cloud Map (PCM) layer remains very limited~\cite{elghazaly2023high}.

To update the PCM layer, change detection provides valuable cues by indicating where modifications have occurred and what types of changes are present. Change detection can operate across multiple modalities, including 2D images~\cite{pollard2007change}, 3D point clouds~\cite{kim2021updating}, and cross-modality approaches that fuse different sensor types~\cite{he2022diff}. These techniques produce change masks or labeled regions that highlight discrepancies between existing map data and new observations. However, simply detecting differences is not sufficient for effective map updating. A significant gap remains in understanding how to robustly integrate detected changes into the existing map, including the addition of new objects and removal of obsolete elements, while maintaining global geometric consistency and structural integrity. This challenge is further amplified by noisy detection outputs or coarse change masks derived from different modalities, particularly from image-based change detection, which may lack precise geometric detail or suffer from alignment errors. Addressing these issues requires developing methods that can reconcile partial observations, resolve ambiguities, and ensure that updates preserve the overall accuracy and usability of the HD map for downstream applications such as localization and planning.

To address this gap, we introduce~\textbf{SceneEdited}, the first large-scale dataset designed specifically for city-scale 3D PCM updating using images to guide changes in LiDAR-based maps. SceneEdited provides realistic urban-scale point cloud scenes with systematically synthesized modifications, creating controlled out-of-date versions that simulate additions, removals, and complex environmental evolution. Unlike prior datasets focused solely on change detection, SceneEdited directly supports both up-to-date and outdated PCM and corresponding up-to-date 2D image and LiDAR scans. In addition, we offer a comprehensive toolkit that allows researchers to automatically generate new edited scenes, enabling precise control over the types and magnitudes of changes for robust training and standardized evaluation.

\textbf{Our key contributions are:}
\begin{itemize}
    \item The first city-scale dataset for 3D point cloud map updating guided by 2D images.
    \item Controlled synthesis of realistic scene changes to create out-of-date map versions.
    \item A toolkit for automatically generating new edited scenes with controllable changes.
    \item Standardized benchmark tasks and metrics for reproducible evaluation.
\end{itemize}

\begin{table*}[htbp]

    \centering
    
    \resizebox{\textwidth}{!}{
        \begin{tabular}{l  c  c  c  c  c  c r}
            \toprule
            
            \textbf{Dataset} & \textbf{Year} & \textbf{Lidar} & \textbf{Images} & \textbf{HD Map} & \textbf{CD} & \textbf{Map Update} & \textbf{Scale}\\
            \midrule
            BD dataset~\cite{yew2021city} & 2021 & - & Y & - & Image & - & 30 image pairs\\
            \midrule
            Urb3DCD v1, 2~\cite{degelis2021change, de2023siamese} & 2021, 23 & Y & - & - & PCM & - \\
            \midrule
            Change3D~\cite{nagy2021changegan} & 2021 & Y & - & - & PCM & - & 22K Lidar scan pairs\\
            \midrule
            SZTAKIBudapest~\cite{zovathi2022point}& 2022 & Y & - & - & PCM & - & 3 street scene pairs\\
            \midrule
            SLPCCD~\cite{wang2023end} & 2023 & Y & - & - & PCM & - & 600+ street-scene pairs\\
            \midrule
            SHREC2023~\cite{gao2023shrec} & 2023 & Y & - & - & PCM & - & 100+ street-scene pairs\\ 
            \midrule
            3DHD CityScenes~\cite{plachetka20223dhd, plachetka2023dnn} & 2022, 23 & Y & - & Y & PCM & - & 30K+ pole-like changed objects\\
            \midrule
            \midrule
            Argoverse~\cite{chang2019argoverse, wilson2023argoverse} & 2019, 21 & Y & Y & Y & - & - & 1K drive logs\\
            \midrule
            Argoverse TbV~\cite{lambert2022trust} & 2021 & Y & Y & Y & Vector Map & - & 1K+ drive logs\\
            \midrule
            \midrule
            \ours{} (ours) & 2025 & Y & Y & Y & PCM \& Image & PCM & \makecell[r]{800+ up-to-date scene\\870M up-to-date voxels\\73 km trajectory\\covering 3$\text{km}^2$ drive lane\\2K outdated scene\\23K+ changed objects\\2M changed voxels}\\
        
            \bottomrule
        \end{tabular}
    }
    \caption{\textbf{\textit{Dataset Comparison:}} We are the first dataset explicitly support map updating. Notably, the HD Map is multi-layer structure containing Point Cloud Map (PCM) and Vector Map.}
    \label{tab:existing-dataset}
\end{table*}

\section{Related Work}
\label{sec:relwork}

\textbf{HD map construction and PCM representations:}
The point cloud map (PCM) forms the geometric backbone of HD maps, encoding ground altitude, lane-level features, and spatial context for semantic annotations~\cite{elghazaly2023high}. PCMs are typically built using high-end mobile mapping platforms equipped with precise GNSS, high-performance IMUs, and multi-modal sensors including LiDAR and high-resolution cameras~\cite{elghazaly2023high, plachetka20223dhd}. Manual ground control point (GCP) measurements further enhance global accuracy by anchoring maps to geodetic frames~\cite{tsai2023alternative}. Dynamic objects are filtered during post-processing to yield a clean, static map suitable for localization. For storage and inference, PCMs may be voxelised or compressed as 3D occupancy grids, while original raw scans are archived for future refinement~\cite{elghazaly2023high}.

\textbf{Change detection in 3D scenes:}
Numerous methods detect scene changes using image-based or multi-modal inputs, often as precursors to map revision. Early work used voxel-based appearance models under consistent lighting~\cite{pollard2007change}, or computed depth-change probabilities from temporal image sequences for SfM updates~\cite{sakurada2013detecting}. More recent approaches rely on dense correspondence matching via COLMAP pipelines~\cite{schoenberger2016sfm, schoenberger2016mvs} or deep networks~\cite{yew2021city}, often projecting multi-view imagery through 3D models to detect inconsistencies via photometric error~\cite{taneja2011image, palazzolo2018fast}. LiDAR–RGB fusion has also been explored for semantic-aware change detection in urban environments~\cite{qin20143d}, while nearest‑neighbour aggregation of geometric features improves efficiency on unstructured LiDAR scans~\cite{wang2023end}. In parallel, LiDAR-only change detection has advanced using 3D ray tracing~\cite{underwood2013explicit}, range-image transformations with CNNs~\cite{nagy2021changegan} and Markov Random Field~\cite{zovathi2022point}, and Siamese architectures on multi-temporal scans~\cite{degelis2021change, de2023siamese, gao2023shrec}. However, most of these approaches stop at change localization, without addressing how to integrate detected changes into an updated map, as shown in \cref{tab:existing-dataset}. 

\textbf{HD map updating and dataset limitations:}
While HD maps are frequently the target of change detection, datasets for their geometric (PCM) layers are rare and often restricted. Prior work has tackled changes in lane-level semantics using image inputs and proprietary Korean datasets~\cite{heo2020hd}, visual odometry for road-marking drift~\cite{pannen2019hd}, or image-segmentation for landmark change detection~\cite{zhang2021real}. TbV~\cite{lambert2022trust}, under Argoverse~\cite{wilson2023argoverse}, enables semantic-layer change detection but excludes point-cloud geometry. A few recent efforts attempt PCM-level updates using multi-vehicle LiDAR fusion~\cite{kim2021updating}, though data is unreleased and modality-limited. Overall, no existing dataset supports controlled, reproducible, and city-scale benchmarking of PCM-level map editing from camera input.

\section{Problem Formulation}
\label{sec:prob}

Given an \emph{outdated} 3-D point-cloud map \(P_{\text{out}}\) and a set of current, geo-referenced RGB images 
\(\mathcal{I}_{\text{curr}}=\{(I_k,T_k)\}_{k=1}^{K}\),  
estimate an \emph{updated} map \(P_{\text{upd}}\) that reproduces the current scene geometry \(P^*_{\text{upd}}\):

\begin{equation}
\begin{aligned}
    P_{\text{upd}} &= \texttt{MapUpdate}\!\bigl(P_{\text{out}}, \mathcal{I}_{\text{curr}}\bigr), \\[4pt]
    \text{with} \qquad P_{\text{upd}} &\approx P^*_{\text{upd}} .
\end{aligned}
\end{equation}

The approximation \(P_{\text{upd}}\approx P^*_{\text{upd}}\) is quantified at evaluation time by geometric distances (e.g.\ Chamfer, Hausdorff) between the estimated and ground-truth point clouds.

This formulation casts \textit{map updating} as a direct transformation that fuses fresh sensor evidence into an outdated map.  Crucially, \textit{explicit change detection is \emph{not} required}: the operator \texttt{MapUpdate} can learn to infer and apply geometry differences implicitly by analysing $P_{\text{out}}$ and $\mathcal{I}_{\text{curr}}$ jointly.

\medskip
\noindent
In many practical pipelines, however, it is advantageous to insert an \textit{explicit change-detection module} that pinpoints where updates are needed.  Let $C_{\text{curr}}$ denote a binary or probabilistic change mask derived from the same inputs.  The update process can then be expressed as
\begin{equation}
\begin{aligned}
    P_{\text{upd}} &= \texttt{MapUpdate}\!\bigl(P_{\text{out}}, \mathcal{I}_{\text{curr}}, C_{\text{curr}}\bigr), \\[2pt]
    C_{\text{curr}} &= \texttt{ChangeDetect}\!\bigl(P_{\text{out}}, \mathcal{I}_{\text{curr}}\bigr).
\end{aligned}
\end{equation}

\noindent
Here, $C_{\text{curr}}$ highlights prospective additions and deletions, enabling the updater to focus computation and often improving both efficiency and accuracy.

\def\gridwidth{0.24\textwidth}
\setlength{\tabcolsep}{1pt}

\begin{figure*}[t]
  \centering
  \begin{tabular}{c c | c c}

    \multicolumn{2}{c|}{$P_{out}$} & \multicolumn{2}{c}{$P^*_{upd}$} \\

    \includegraphics[width=\gridwidth]{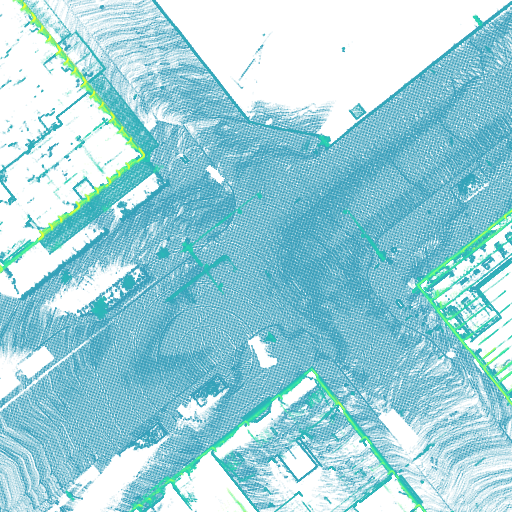} & 
    \includegraphics[width=\gridwidth]{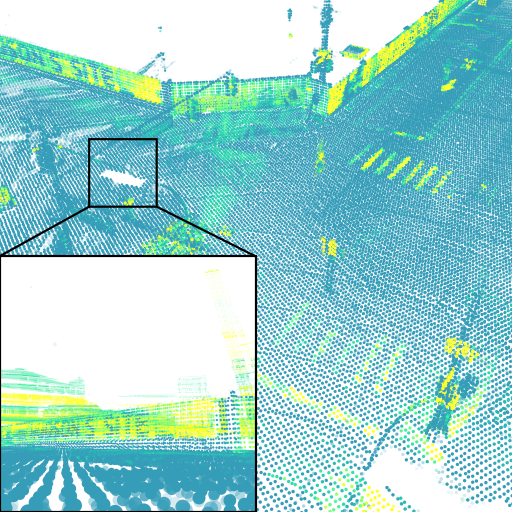} &
    \includegraphics[width=\gridwidth]{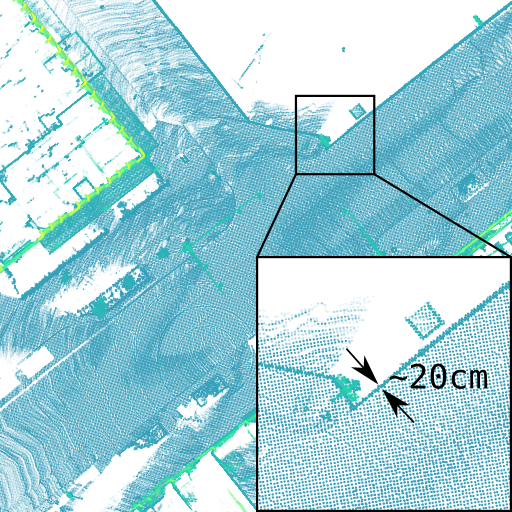} &
    \includegraphics[width=\gridwidth]{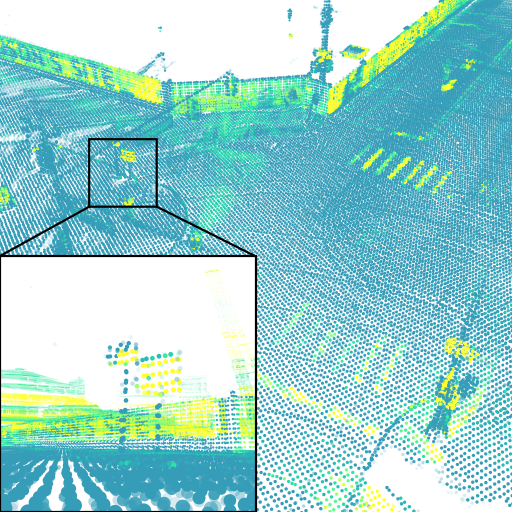} \\

  \end{tabular}
  
  \caption{\textbf{\ours{} – Outdated and Updated Scene PCM Visualization:} We compare the outdated PCM $P_{\text{out}}$ and the ground truth updated PCM $P^*_{\text{upd}}$ using both top-down and third-person views. The top-down view is rendered with orthographic projection, while the third-person view uses perspective projection. In the top-down view, point clouds are colored by altitude to highlight building structures such as walls. We also demonstrate the PCM's geometric accuracy by measuring wall thickness. In the third-person view, point clouds are rendered using LiDAR intensity, where road markings typically appear with higher intensity values. This makes lane markings clearly distinguishable in the visualization.}

  \label{fig:sample-data}
\end{figure*}
\def\gridwidth{0.16\textwidth}
\setlength{\tabcolsep}{1pt}

\begin{figure*}
  \centering
  \begin{subfigure}{\gridwidth}
    \includegraphics[width=\linewidth]{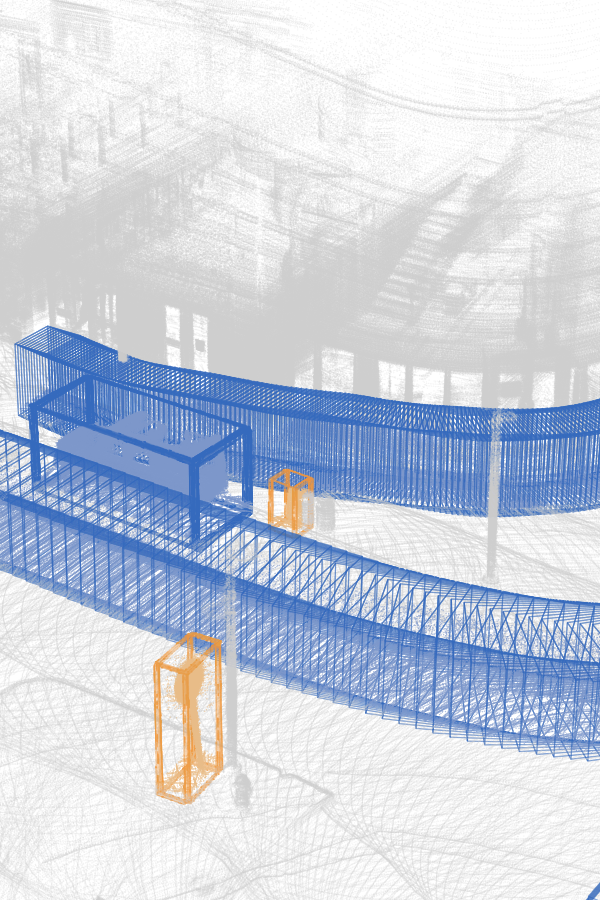}
    \caption{}
    \label{fig:p_a}
  \end{subfigure}
  \hfill
  \begin{subfigure}{\gridwidth}
    \includegraphics[width=\linewidth]{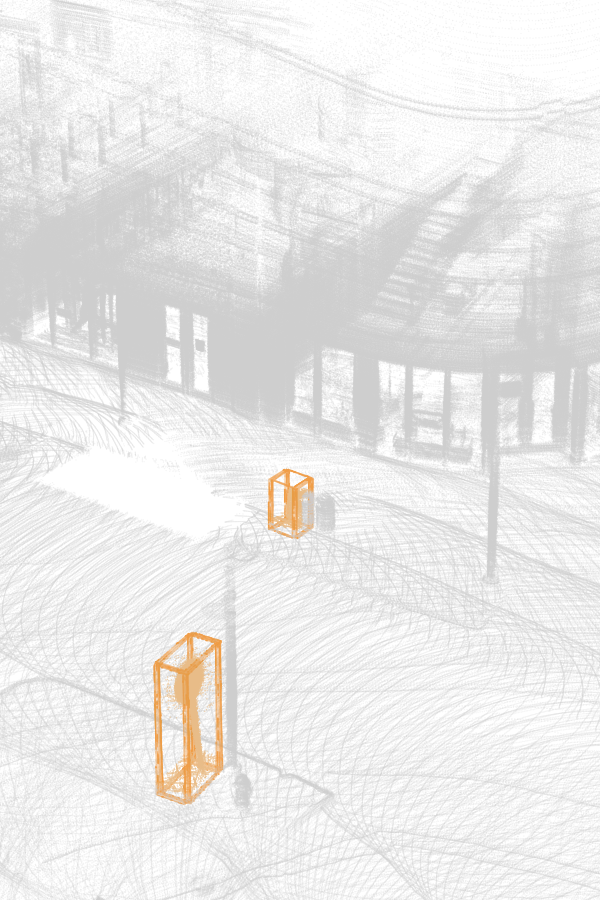} 
    \caption{}
    \label{fig:p_b}
  \end{subfigure}
  \hfill
  \begin{subfigure}{\gridwidth}
    \includegraphics[width=\linewidth]{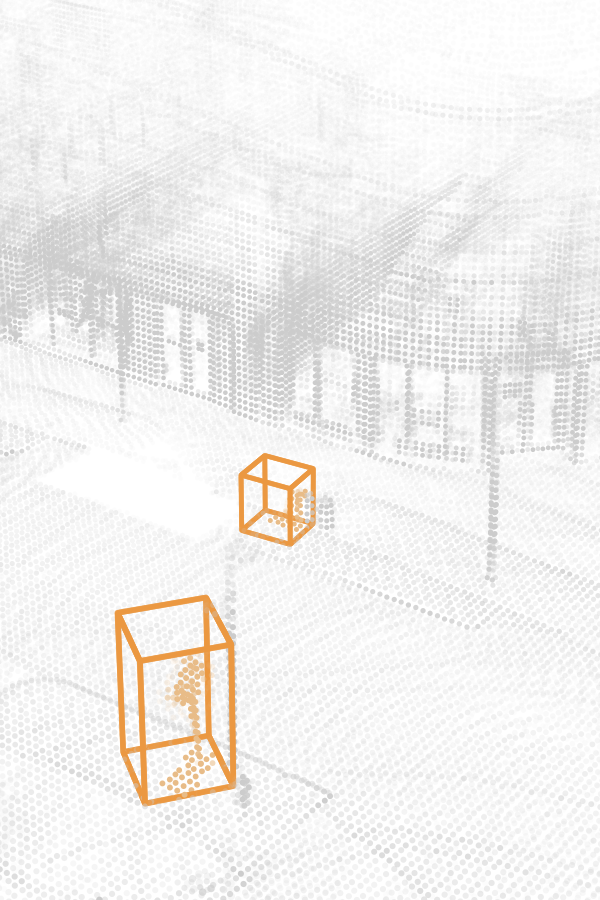} 
    \caption{}
    \label{fig:p_c}
  \end{subfigure}
  \hfill
  \begin{subfigure}{\gridwidth}
    \includegraphics[width=\linewidth]{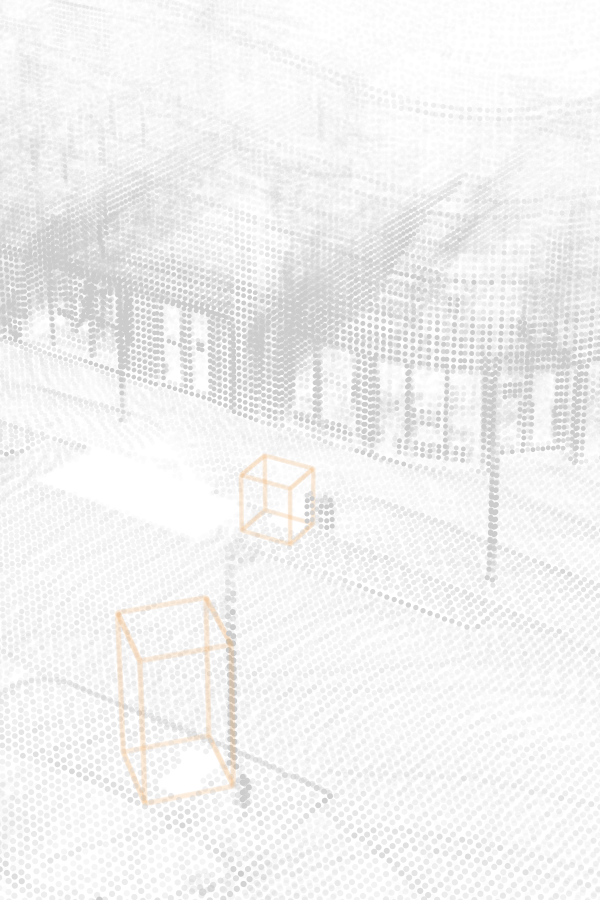} 
    \caption{}
    \label{fig:p_d}
  \end{subfigure}
  \hfill
  \begin{subfigure}{\gridwidth}
    \includegraphics[width=\linewidth]{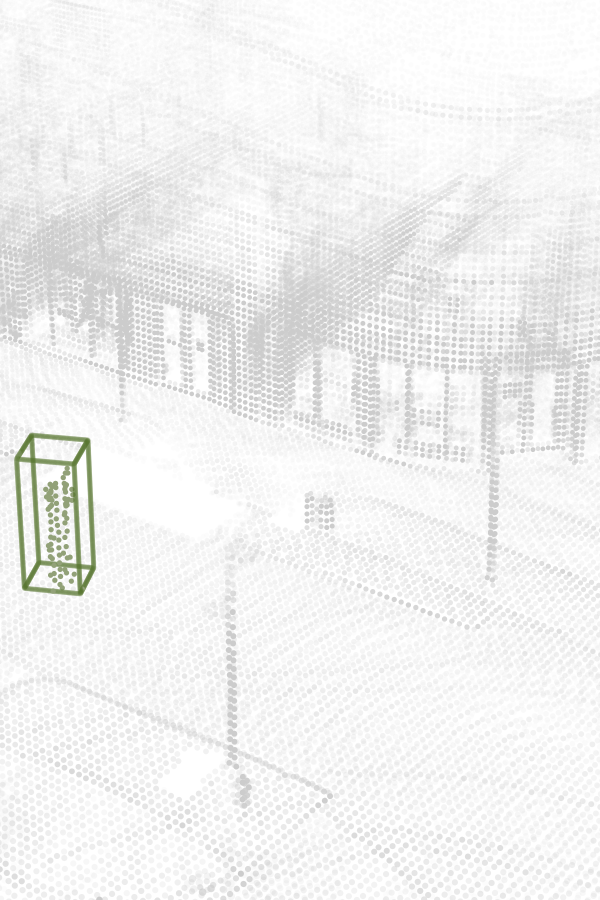} 
    \caption{}
    \label{fig:p_e}
  \end{subfigure}
  \hfill
  \begin{subfigure}{\gridwidth}
    \includegraphics[width=\linewidth]{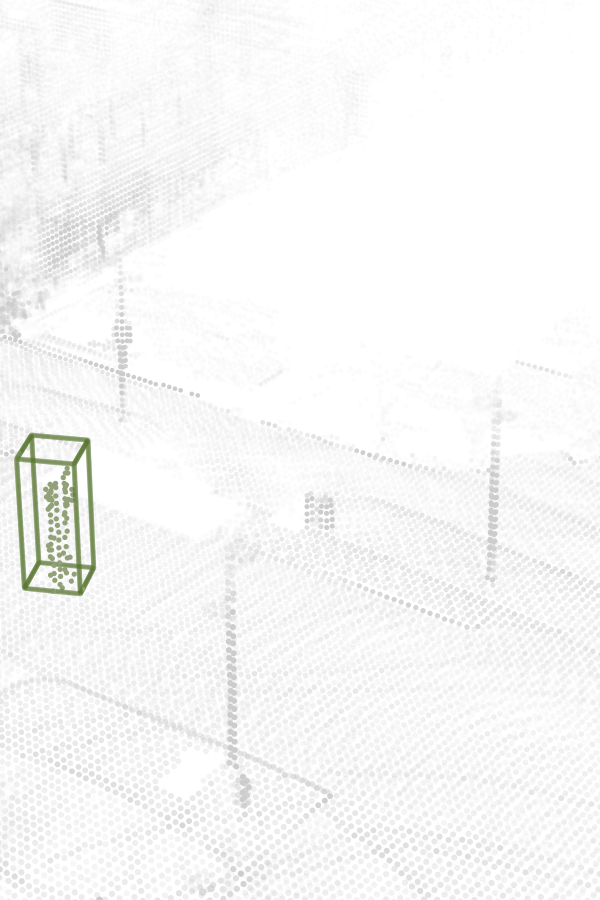} 
    \caption{}
    \label{fig:p_f}
  \end{subfigure}

  \caption{Overview of the processing pipeline from raw LiDAR scans to $P^*_{\text{upd}}$ and $P_{\text{out}}$. 
(a) Raw LiDAR points with dynamic cuboids (blue) and static cuboids (orange). 
(b) LiDAR points after removing dynamic objects. 
(c) Voxelized static points ($P^*_{\text{upd}}$). 
(d--f) Examples of outdated scenes ($P_{\text{out}}$): removing static objects, inserting new objects (green), and manually removing buildings, respectively.
Note that the orientation of bounding boxes changes between (b) and (c) because we recompute their orientation using the principal eigenvector of the points inside each box. These bounding boxes are used solely for quantization and visualization purposes.
}
  \label{fig:create-pipeline}
\end{figure*}

\def\gridwidth{0.24\textwidth}
\setlength{\tabcolsep}{1pt}

\begin{figure*}[t]
  \centering
  \begin{tabular}{c c c c }

    & \multicolumn{3}{c}{depth image} \\
    $\mathcal{I}_{\text{curr}}$ + $C_{\text{curr}}$ & $\mathcal{L}_{c}$ & $P_{out}$ & $P^*_{upd}$ \\
     

    \includegraphics[width=\gridwidth]{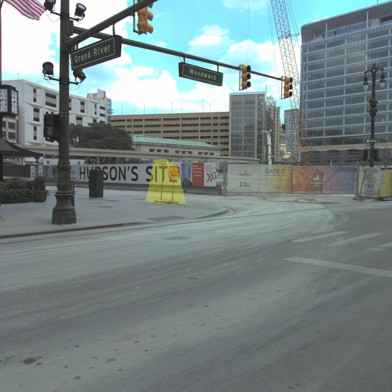} & 
    \includegraphics[width=\gridwidth]{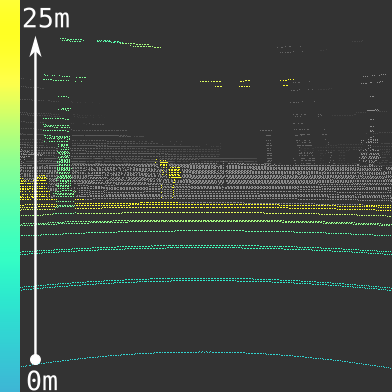} & 
    \includegraphics[width=\gridwidth]{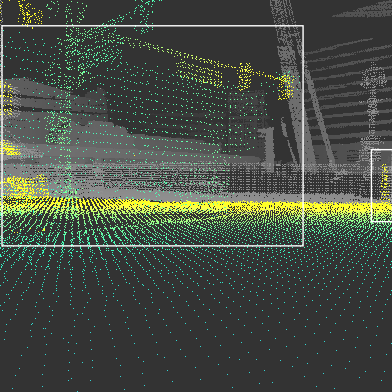} & 
    \includegraphics[width=\gridwidth]{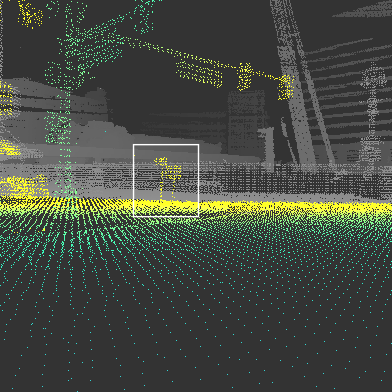} \\

  \end{tabular}
  
  \caption{\textbf{\ours{} – Current Urban Scene and Depth Images:} We visualize the current urban scene image $\mathcal{I}_c$ alongside the corresponding depth images generated from the current LiDAR $\mathcal{L}_c$, the outdated PCM $P_{\text{out}}$, and the ground truth updated PCM $P^*_{\text{upd}}$. To enhance the visibility of the sparse depth images, we apply color rendering within a 25-meter range. Bounding boxes are overlaid on the depth images of $P_{\text{out}}$ and $P^*_{\text{upd}}$ to indicate obsolete and missing objects with respect to $P_{\text{out}}$. The 2D pixel-level change map $C_{\text{curr}}$ is also shown on the image.}

  \label{fig:sample-data2}
\end{figure*}

\setlength{\tabcolsep}{6pt}
\begin{table}[htbp]

    \centering
    
        \begin{tabular}{l | r | r | c }

            \toprule
            
            name & quantity & space (GB) & portable\\
            \midrule
            \textbf{Source Dataset:} & & & \\ 
            Argoverse & 850 & 900 & - \\
            \midrule
            \textbf{SceneEdited:} & & & \\
            inter. of $P^*_{\text{upd}}$ & 847 & 42.69 & V \\ 
            inter. of $P_{\text{out}}$ & 2255 & 0.13 & V \\
            inter. of $\mathcal{I}_{\text{curr}}$ & 847 & 0.01 & V \\
            $P^*_{\text{upd}}$ & 847 & 39.95 & - \\
            $P_{\text{out}}$ & 2255 & 108.43 & - \\
            $\mathcal{I}_{\text{curr}}$, $C_{\text{curr}}$ & 487K & 368.92 & - \\
            
            \bottomrule
        \end{tabular}

    \caption{\textbf{\textit{Portability:}} The portable dataset stores only essential components required for distribution. Using the Scene Editing Toolkit, the full dataset can be losslessly reconstructed from this compact representation.}
    \label{tab:table-portability}
\end{table}
\section{SceneEdited Dataset}
\label{sec:dataset}

\ours{} provide both updated and outdated 3D scenes, detailed change masks, and aligned images to enable supervised learning on change detection and map updating. Notably, while we include LiDAR scans as auxiliary data, they are intended to assist rather than serve as the primary resource for updating the map.

\subsection{Up-To-Date Scene}

Each sample in \ours{} includes an up-to-date 3D point cloud that serves as the ground-truth representation of the current urban environment. The dataset comprises over 800 such up-to-date scenes, constructed using Argoverse’s high-quality LiDAR scans with accurate calibration and city-scale coverage. Notably, these scans are from VLP-32C LiDAR sensors which are different from traditional surveying and mapping LiDAR scanner~\cite{elghazaly2023high}. But we observed that the resulting quality is comparable, as shown in~\cref{fig:sample-data}.

To ensure that the resulting PCM are suitable for HD map maintenance, we remove all dynamic objects from the raw scans, ensuring each scene consists only of static objects (see \cref{fig:p_a} \& \cref{fig:p_b} for an illustration of this pipeline). Specifically, we leverage the 3D cuboid annotations provided by Argoverse to identify and filter out points associated with dynamic objects such as vehicles, cyclists, and pedestrians. A full list of dynamic objects is in the Supplement (Sec.~A).

After isolating the static, dense LiDAR points, we further voxelize them to produce $P_{\text{upd}}$ in a format consistent with common PCM scene representations, using a 20 cm resolution. This voxelization step yields consistent, noise-reduced static voxel point clouds that serve as the canonical up-to-date maps. Additionally, we store customized intermediate results of $P_{\text{upd}}$ to enable tracking of changes from voxel space to the original raw LiDAR scan points.

\subsection{Outdated Scene}

\ours{} generates outdated versions of up-to-date scenes to simulate realistic urban change scenarios for HD map maintenance research. All edits are performed on the voxelized static point clouds described in the previous section, ensuring precise and easily trackable changes at voxel-level resolution.

To create these outdated scenes, we employ both automated and manual editing strategies. We first use the static cuboid annotations provided by Argoverse to selectively remove static objects from the scenes. This systematic removal simulates typical urban changes, such as missing infrastructure or demolished elements, in a reproducible manner (see \cref{fig:p_c} \& \cref{fig:p_d}). Additionally, we manually remove points corresponding to overpasses, trees, and buildings in selected scenes to introduce greater variability and increase the challenge of the updating task. This process is carried out by directly selecting voxels and recording their corresponding indices for removal (see \cref{fig:p_e} \& \cref{fig:p_f}). A full list of static objects by Argoverse is in the Supplement (Sec.~A).

To further enrich the diversity of changes, we construct a patch database containing over 1,000 static object point clouds. These patches are automatically inserted into scenes as occlusion-free change objects using the ground plane information from Argoverse, ensuring plausible and realistic placement within urban contexts (see \cref{fig:p_d} \& \cref{fig:p_e}). A summary of the types and counts of introduced change objects is provided in the Supplement (Sec.~B).

Finally, we store intermediate results of $P_{\text{out}}$ to explicitly track changed objects and their locations. This enables consistent labeling and facilitates further processing, such as supervised learning of change detection and map updating models.

\subsection{Sensor Data and Change Map}

\ours{} uses the raw sensor measurements from the Argoverse dataset, including high-resolution LiDAR scans and RGB images, along with the provided intrinsic and extrinsic calibration parameters. These calibrated, synchronized sensor modalities offer a realistic foundation for simulating city-scale HD map maintenance tasks.

To transfer change information from the 3D outdated scene to these sensor measurements, we project points from $P_{\text{out}}$ onto each image using the known calibration. We then compare projected change points with synchronized LiDAR scans to remove occluded or spurious detections arising from dynamic objects. Finally, we compute the convex hull of the filtered projected change points to generate precise image-space change labels for each object, as shown in \cref{fig:sample-data2}. A step by step visualizations of how change map is inferred is in the Supplement (Sec.~C).

In addition to these image-space labels, we provide change maps for both raw LiDAR scans and voxelized scenes by combining intermediate results of $P^*_{\text{upd}}$ and $P_{\text{out}}$. This comprehensive labeling enables cross-modality map updating by providing consistent, aligned change supervision across 2D images and 3D point clouds.

\subsection{Automatic Scene Editing Toolkit}

\ours{} includes an automatic scene editing toolkit that enables efficient storage, customization, and extensibility of the dataset. By storing compact intermediate representations of both $P^*_{\text{upd}}$ and $P_{\text{out}}$, the toolkit can export a highly portable version of the dataset that reduces storage requirements by over 90\% while still supporting exact reconstruction of the original edited scenes. The detailed storage savings are summarized in \cref{tab:table-portability}. This approach ensures the practical distribution and usability of large-scale benchmarks.

The toolkit also ensures the trackability of all changes through intermediate results, enabling users to precisely identify and recover specific additions or deletions. Additionally, it supports dividing and unifying individual object changes across multiple outdated scenes to facilitate targeted unit testing. Users can also create new edited versions of scenes by programmatically removing existing objects or adding new patches from the provided database, enabling systematic and scalable expansion and adaptation for diverse research needs.

\begin{figure*}[ht]
  \centering
  \includegraphics[width=\textwidth]{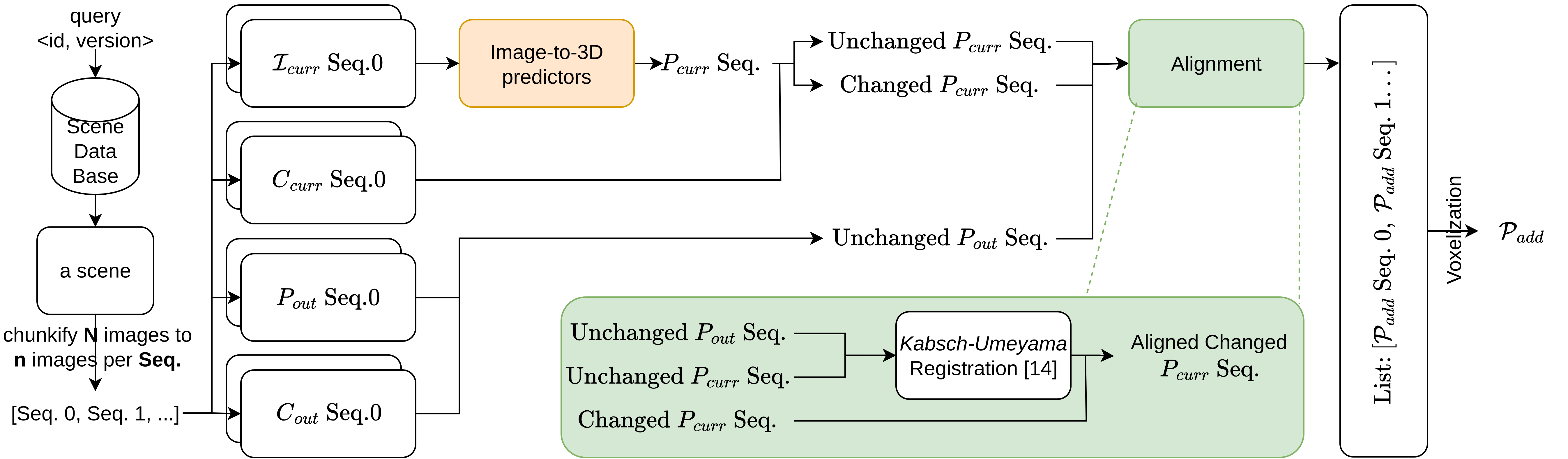}
  \caption{Overview of the point addition from image-to-3D predictor pipeline from querying a scene to $P_{\text{add}}$.}
  \label{fig:pipeline}
\end{figure*}
\def\gridwidth{0.23\textwidth}
\setlength{\tabcolsep}{1pt}

\begin{figure*}[t]
  \centering
  \begin{tabular}{c | c c c }

     & VGGT~\cite{wang2025vggt} & MASt3R~\cite{leroy2024grounding} & DUSt3R~\cite{wang2024dust3r} \\

    \toprule

    $\mathcal{I}_{\text{curr}}$ + $C_{\text{curr}}$ & \multicolumn{3}{c}{Top-down view of registration between Image points (yellow) and $P_{\text{out}}$ (Blue)} \\
    

    \includegraphics[width=\gridwidth]{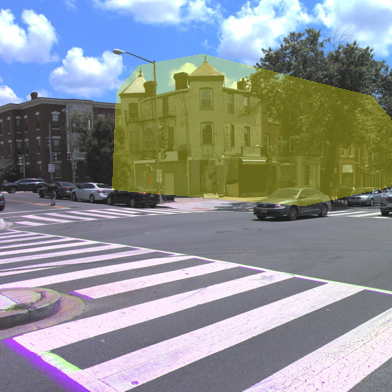} & 
    \includegraphics[width=\gridwidth]{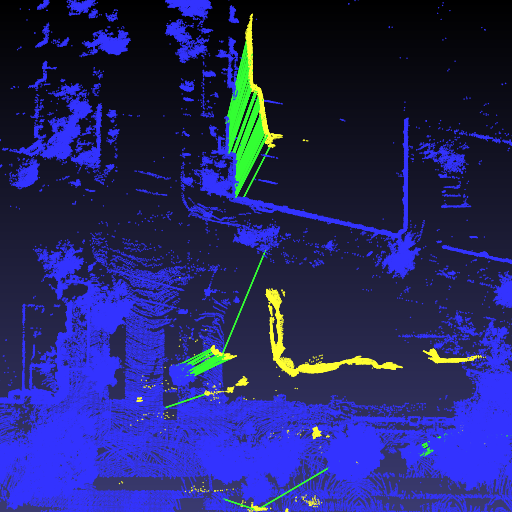} & 
    \includegraphics[width=\gridwidth]{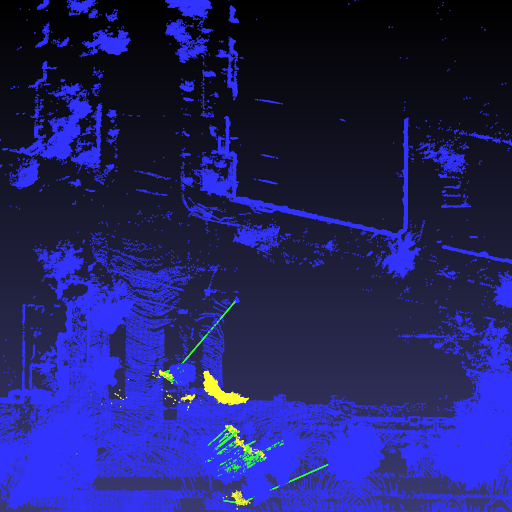} & 
    \includegraphics[width=\gridwidth]{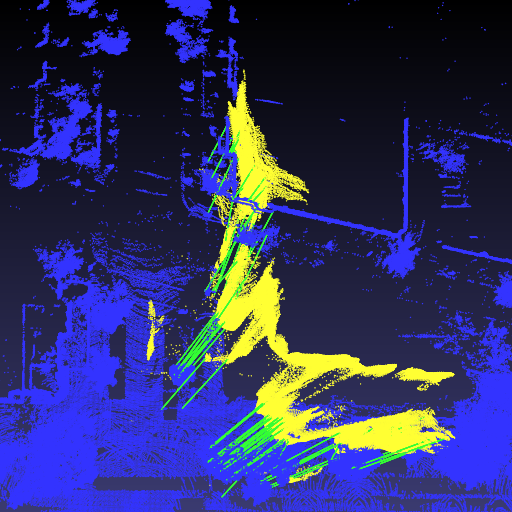} \\
    
    GT $P^*_{add}$ & \multicolumn{3}{c}{Image Point Addition Point $P_{add}$ (yellow) and $P^*_{add}$ (Blue) Comparison} \\
    
    \includegraphics[width=\gridwidth]{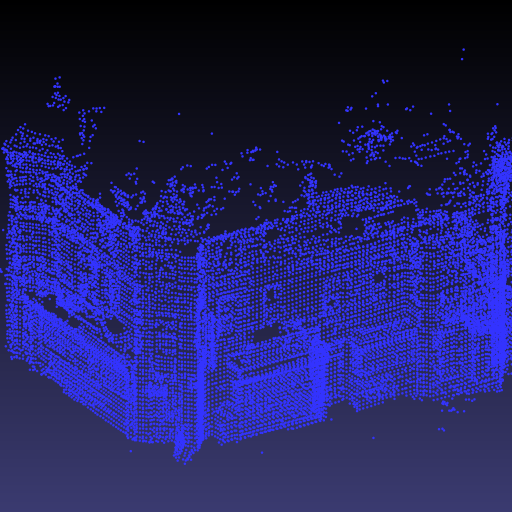} & 
    \includegraphics[width=\gridwidth]{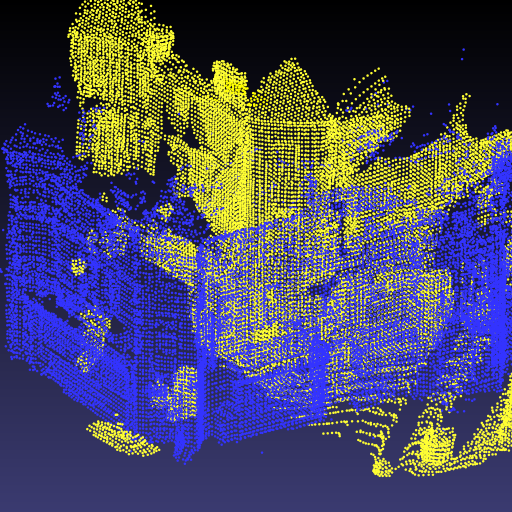} & 
    \includegraphics[width=\gridwidth]{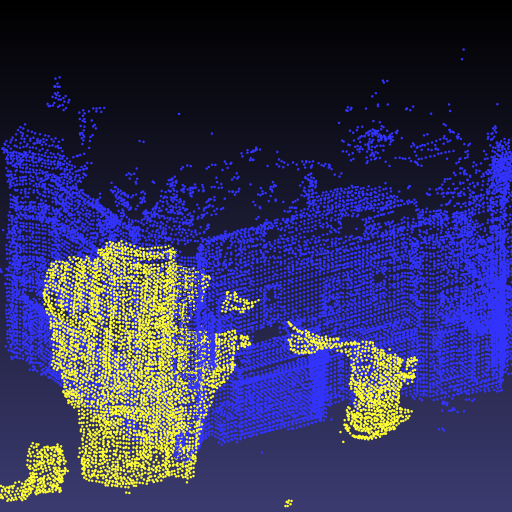} & 
    \includegraphics[width=\gridwidth]{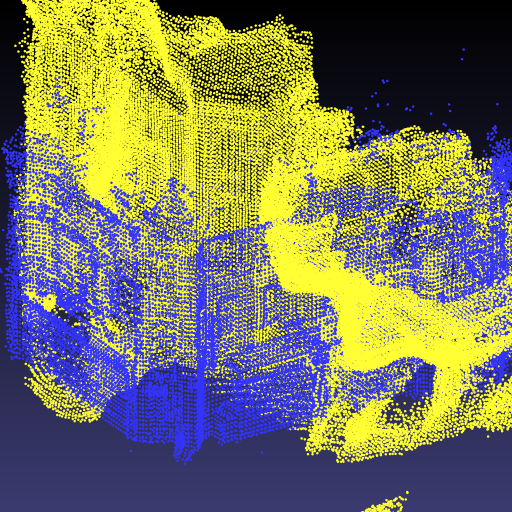} \\

  \end{tabular}
  \caption{\textbf{Point Addition from Image:} We show registration results by comparing the outdated PCM $P_{\text{out}}$ with image-derived points in the first row. Green lines highlight several matched points, providing visual cues for registration quality. In the second row, we compare the missing building structure in $P_{\text{out}}$ with the image-derived points overlaid on the image change map to assess reconstruction quality. The results show that MASt3R~\cite{leroy2024grounding} registers accurately to the scene but fails to recover the complete structure of the missing building, while VGGT~\cite{wang2025vggt} reconstructs the full structure well but suffers from registration misalignment.}

  \label{fig:experiment}
\end{figure*}
\def\gridwidth{0.12\textwidth}
\setlength{\tabcolsep}{1pt}

\begin{figure*}
  \centering
  \includegraphics[width=\textwidth]{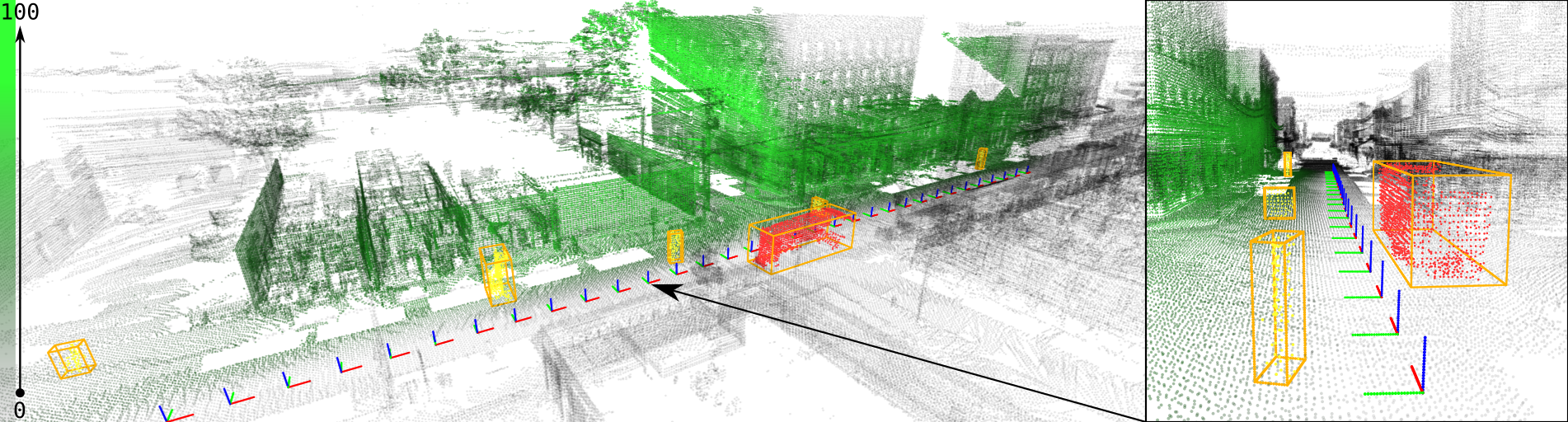}
  \caption{\textbf{Visibility from a Single Camera:} We visualize point visibility by accumulating the number of images in which each scene point is observed. For obsolete objects (orange bounding boxes), visible points are shown in yellow, while invisible points are shown in red. Since the selected camera is mounted on the left side of the vehicle, the red change object located on the right side of the car trajectory is clearly outside the camera's field of view.}
  \label{fig:visibility}
\end{figure*}

\section{Evaluation}
\label{sec:evaluation}

This task requires generating an updated 3D map that accurately represents the environment by giving an outdated 3D map $P_{\text{out}}$ and current RGB observations $\mathcal{I}_{curr}$. The goal is to maintain map fidelity while correctly incorporating new structures and removing outdated elements to produce a change-free 3d map $P_{upd}$. As the change part can only take very small portion in the entire scene, we divide the $P_{\text{add}}$, $P_{\text{del}}$, $P^*_{\text{add}}$ and $P^*_{\text{del}}$ by the following set operation for evaluation.

\begin{equation}
\begin{alignedat}{2}
P^*_{\text{add}} &= P^*_{\text{upd}} - P_{\text{out}} \quad &\qquad 
P^*_{\text{del}} &= P_{\text{out}} - P^*_{\text{upd}} \\
P_{\text{add}}   &= P_{\text{upd}} - P_{\text{out}} \quad &\qquad
P_{\text{del}}   &= P_{\text{out}} - P_{\text{upd}}
\end{alignedat}
\end{equation}
  
To assess the \emph{geometric similarity} between predicted and ground-truth maps, we adopt multiple complementary distance-based metrics: \textit{a)} Chamfer Distance ($D_{C}$), which captures average bidirectional alignment over many points~\cite{wu2021density}; \textit{b)} Hausdorff Distance ($D_{H}$), which measures worst-case deviation; \textit{c)} Modified Hausdorff Distance ($D_{MH}$), which accounts for outlier robustness; and \textit{d)} Median Point Distance ($D_{MD}$) to focus on typical errors while ignoring extrema~\cite{zovathi2022point}. These metrics together provide a comprehensive assessment of the update quality.

{\footnotesize
\begin{equation}
D_{C}(P, P^*) = \frac{1}{|P|} \sum_{p \in P} d(p, P^*)^2 + \frac{1}{|P'|} \sum_{p^* \in P^*} d(p^*, P)^2
\label{eq:chamfer}
\end{equation}
}

{\footnotesize
\begin{equation}
D_{H}(P, P^*) = \max \left\{ \max_{p \in P} d(p, P^*), \max_{p^* \in P^*} d(p^*, P) \right\}
\label{eq:hd}
\end{equation}
}

{\footnotesize
\begin{equation}
D_{MH}(P, P^*) = \max \left\{ \frac{1}{|P|} \sum_{p \in P} d(p, P^*), \frac{1}{|P^*|} \sum_{p^* \in P^*} d(p^*, P) \right\}
\label{eq:mhd}
\end{equation}
}

{\footnotesize
\begin{equation}
D_{MP}(P, P^*) = \max \left\{ \underset{p \in P}{\text{median}}\ d(p, P^*), \underset{p^* \in P^*}{\text{median}}\ d(p^*, P) \right\}
\label{eq:mpd}
\end{equation}
}

,where $d(a, B) = \inf_{b \in B} \|a - b\|$.

\setlength{\tabcolsep}{6pt}

\begin{table}[htbp]
\centering
\resizebox{\linewidth}{!}{
    \begin{tabular}{l | c | c | c | c | c | c }

        \toprule
        method & $D_C$  & $D_H$ & $D_{MH}$ & $D_{MP}$ & Runtime \\
            & ($\text{m}^2$) & (m) & (m) & (m) & (s) &fail/all \\
        \midrule
        DUSt3R\cite{wang2024dust3r}     & 565.62 & 42.91 & 13.83 & 11.66 & 177.26 & 06/239 \\
        MASt3R\cite{leroy2024grounding} & 717.95 & 48.41 & 14.37 & 11.33 & 276.49 & 11/239 \\
        VGGT\cite{wang2025vggt}         & 700.29 & 43.99 & 13.94 & 10.98 & 031.10 & 14/239 \\
        
        \bottomrule
        
    \end{tabular}
}
\caption{\textbf{\textit{Point Addition Comparison:}} We compare different feed-forward image-to-3D predictors across 239 test scenes, each containing at least one missing object that needs to be reconstructed. Notably, all distance metrics are averaged over 218 scenes on which all methods succeed.}
\label{tab:addition-benchmark}
\end{table}
\setlength{\tabcolsep}{6pt}
\begin{table}[htbp]
\centering
    \begin{tabular}{c | c | c | c | c }
        \toprule
         $D_C$  & $D_H$ & $D_{MH}$ & $D_{MP}$ & \\
         ($\text{m}^2$) & (m) & (m) & (m) & fail/all \\
        
        \midrule
        
        0.31 & 0.48 & 0.04 & 0.03 & 0/264\\
        
        \bottomrule
    \end{tabular}
\caption{\textbf{\textit{Visibility:}} We estimate the lowest achievable metric value using ground-truth change masks across 264 test scenes, each containing at least one obsolete object that needs to be deleted.}
\label{tab:deletion-benchmark}
\end{table}

\section{Experiments and Results}
\label{sec:approach}

In this section, we evaluate the potential for image-based PCM updating by leveraging ground-truth change masks to isolate the impact of geometric reconstruction and registration quality, independent of change detection errors. We divide map updating into two subproblems: point addition (\cref{subsec:add}) and deletion (\cref{subsec:delete}). For both, we analyze challenges even under ideal conditions, using one camera view and its corresponding ground-truth change map to explore practical limitations.

\subsection{Point Addition}
\label{subsec:add}

Given $P_{\text{out}}$, $\mathcal{I}_{\text{curr}}$, camera parameters, and a change map, the point addition operation aims to generate accurate new 3D points from image-derived change regions. Unlike directly adding new points from LiDAR, this approach is more challenging due to the lack of absolute depth in images and the need to fuse data across modalities. Nevertheless, as image-to-3D predictor methods improve, fusing image-based and LiDAR-based maps for PCM updating has become increasingly feasible.

We evaluate three feed-forward image-to-3D predictors: VGGT~\cite{wang2025vggt}, MASt3R~\cite{leroy2024grounding}, and DUSt3R~\cite{wang2024dust3r} to generate 3D point candidates from image pixels within the change regions. Each predictor estimates camera poses, depth maps, and dense 3D point clouds. However, these reconstructions are not in a globally consistent metric frame and must be registered to $P_{\text{out}}$. To preserve geometric alignment, we use a rigid transformation and employ the Kabsch-Umeyama algorithm \cite{lawrence2019purely} to scale and align unchanged image points to their corresponding $P_{\text{out}}$ regions. Notably, we use ground-truth change maps and camera intrinsics/extrinsics to extract pixel-aligned correspondences for registration. 

To obtain $P_{\text{add}}$, we input five images at a time into the predictors and accumulate the resulting 3D reconstructions within the change masks $C_{\text{curr}}$. These partial reconstructions are then integrated into the updated PCM. We illustrate this process in \cref{fig:pipeline}. Results with larger image sets are provided in the Supplement (Sec.~D). Note that all experiments are computed by single NVIDIA A100 GPU.

Table \ref{tab:addition-benchmark} compares these methods quantitatively, evaluating reconstruction quality and alignment error relative to $P_{\text{add}}$. All experiments use ground-truth change labels to isolate reconstruction accuracy from segmentation errors. We also include qualitative alignment visualizations in \cref{fig:experiment}. The results show that while these methods produce high-quality alignments across frames, residual misalignments remain when registering to $P_{\text{out}}$. This limitation is compounded by dynamic objects present in $\mathcal{I}_{\text{curr}}$, which are not distinguished in the static change map, leading to unreliable registration for dynamic regions.

\subsection{Point Deletion}
\label{subsec:delete}

In contrast to the point addition (\cref{subsec:add}), the point deletion focuses on removing points from $P_{\text{out}}$ that should no longer be part of the updated PCM because they are no longer present in the scene. In these experiments, we explore deletion using ground-truth change masks and $\mathcal{I}_{\text{curr}}$ to identify regions of change.

A key limitation of image-based point deletion is that it can only operate within the camera’s field of view (FOV): only points observed in $\mathcal{I}_{\text{curr}}$ can be confidently evaluated for removal. Points outside the camera's FOV remain unobserved and must conservatively remain in the map, regardless of any change labels, since their absence in the new image provides no evidence they were actually removed from the scene. Notably, the points occluded by other foreground object points should remain as well as they cannot be spotted in $\mathcal{I}_{\text{curr}}$ even in field of view. Based on these rule, we perform visibility experiment to estimate $P_{\text{del}}$ using ground truth change masks.

\cref{tab:deletion-benchmark} reports the evaluation results between the predicted deletions $P_{\text{del}}$ and the ground-truth deletions $P^*_{\text{del}}$. Example scenes are visualized in \cref{fig:visibility}. From both the quantitative and qualitative results, we observe that most obsolete objects are at least partially visible to a single camera. However, visibility alone does not guarantee complete observability. For instance, in the case of a demolished tall building, the upper structure points may not be visible due to the camera’s limited vertical coverage, as visibility tends to decrease with increasing altitude, as illustrated in \cref{fig:visibility}.

\section{Conclusion}
\label{sec:conc}

Due to the shortage of publicly available datasets for 3D point cloud map updating, we not only leverage existing self-driving datasets but also synthesize the new SceneEdited dataset to support map updating research. In contrast to prior works that often focus solely on change detection, our approach targets updating 3D point cloud maps using images, which is both more challenging and better suited to crowdsourcing applications, given that images are far easier to collect at scale than LiDAR point clouds. Moreover, we analyze the key challenges involved in using images to update point clouds, including reconstruction accuracy, registration, and modality fusion. We also release an open-source toolkit and an efficient data format designed to replicate results easily and support the creation of customized outdated scene variants. Finally, our framework can be readily adapted to other self-driving datasets, facilitating broader exploration of multi-sensor fusion and map updating strategies.

{
    \balance
    \small
    \bibliographystyle{ieeenat_fullname}
    \bibliography{main}
}

\clearpage
\maketitlesupplementary

\appendix

\begin{figure*}[h]
  \centering
  \includegraphics[width=0.98\textwidth]{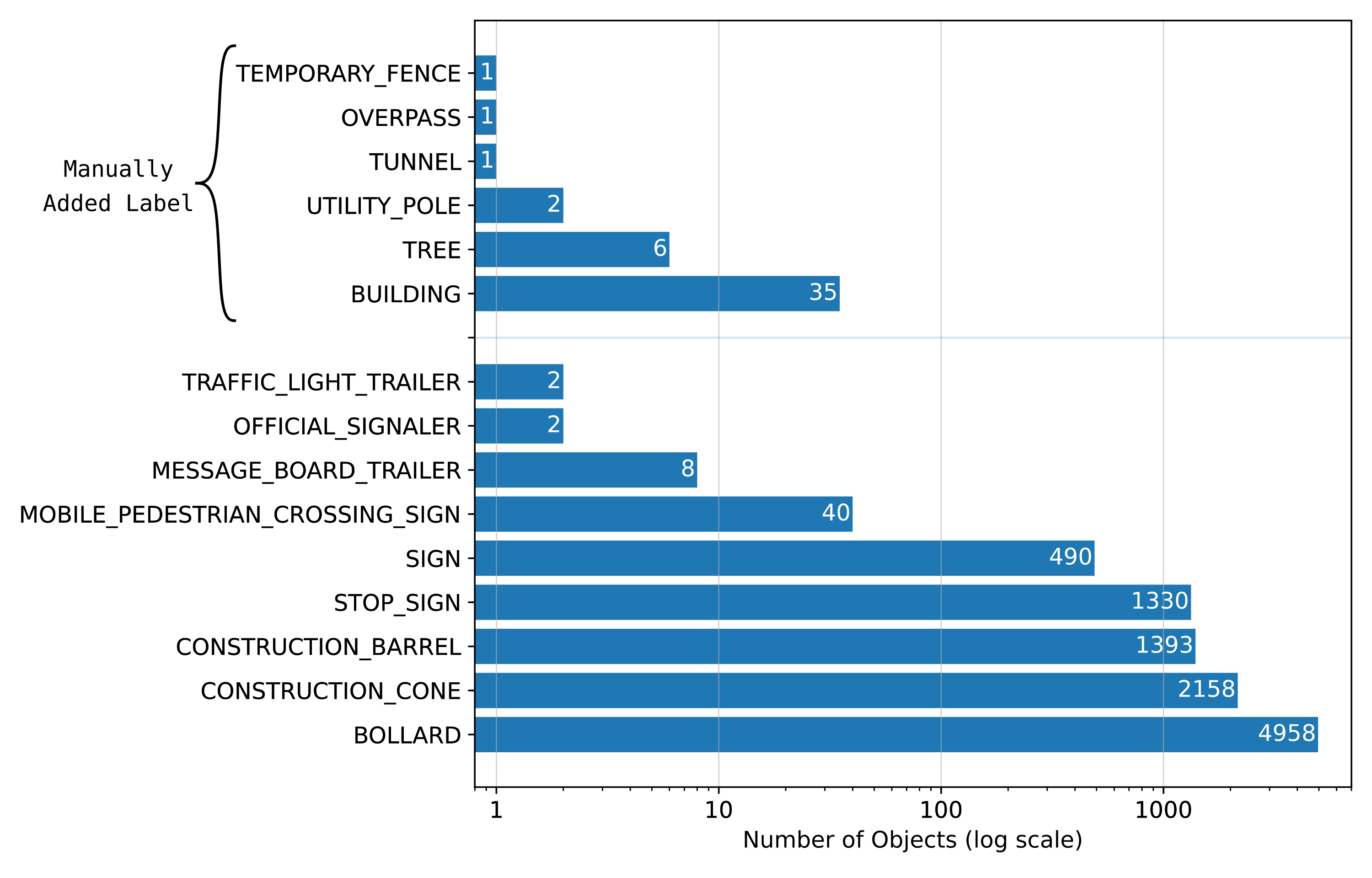}
  \caption{Statistic of missing objects from $P_{out}$ in \ours{} dataset. Each object will remove a number of voxels (points) from $P^*_{upd}$.}
  \label{fig:supp-del}
\end{figure*}

\begin{figure*}[h]
  \centering
  \includegraphics[width=0.98\textwidth]{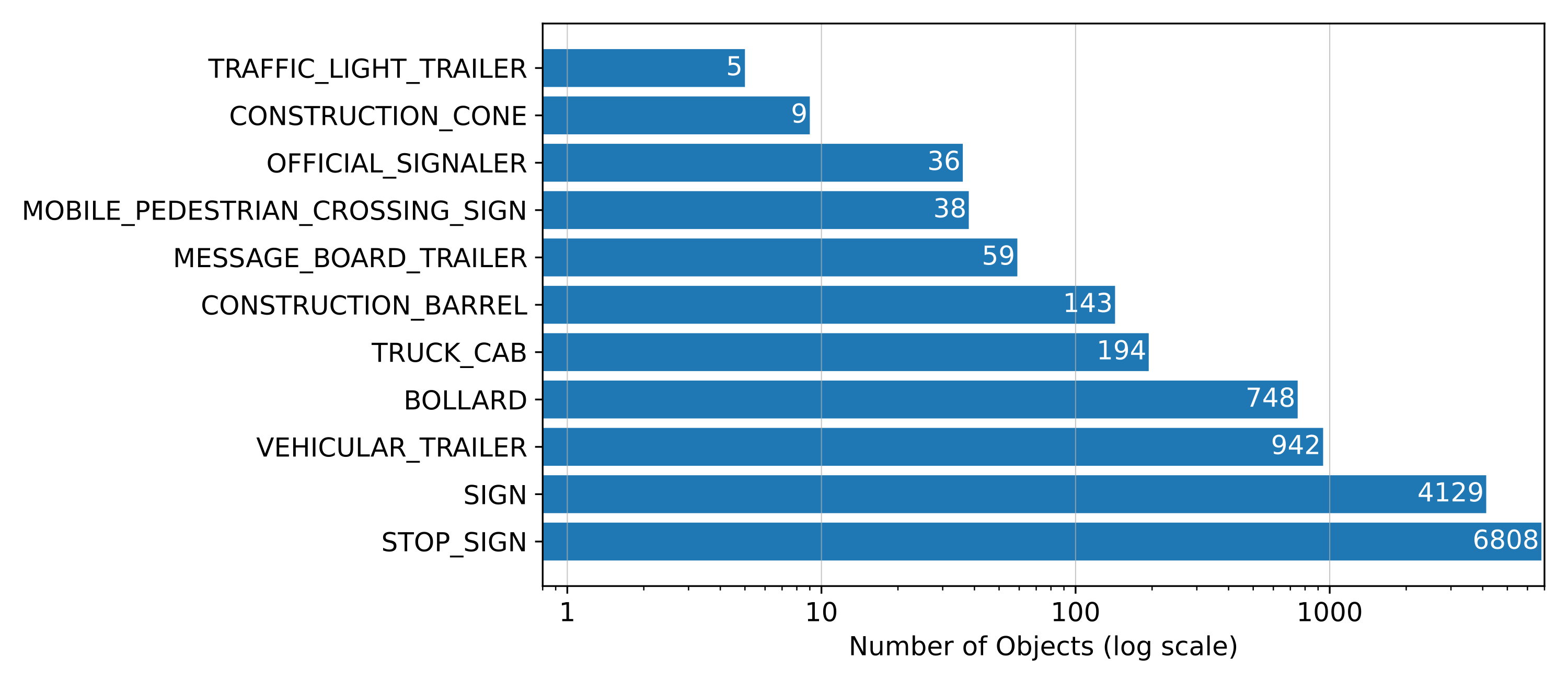}
  \caption{Statistic of outdated objects from $P_{out}$ in \ours{} dataset. Each object will add a number of voxels (points) into $P^*_{upd}$.}
  \label{fig:supp-add}
\end{figure*}

\def\gridwidth{0.19\textwidth}
\def\imgwidth{\linewidth}

\begin{figure*}[h]
  \centering

  \begin{subfigure}{\gridwidth}
    \includegraphics[width=\imgwidth]{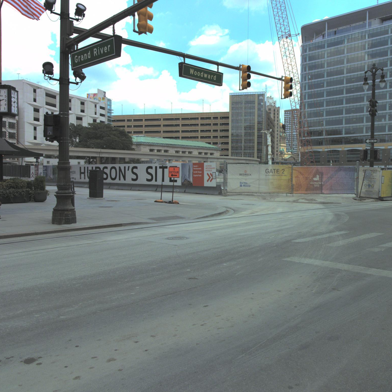}
    \caption{}
    \label{fig:supp-p_a}
  \end{subfigure}
  \hfill
  \begin{subfigure}{\gridwidth}
    \includegraphics[width=\imgwidth]{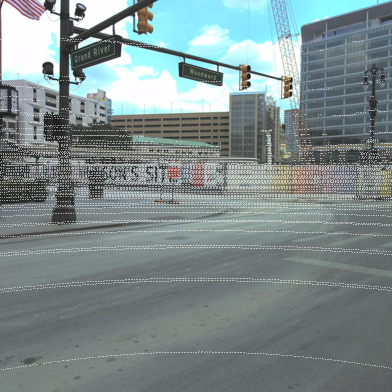}
    \caption{}
    \label{fig:supp-p_b}
  \end{subfigure}
  \hfill
  \begin{subfigure}{\gridwidth}
    \includegraphics[width=\imgwidth]{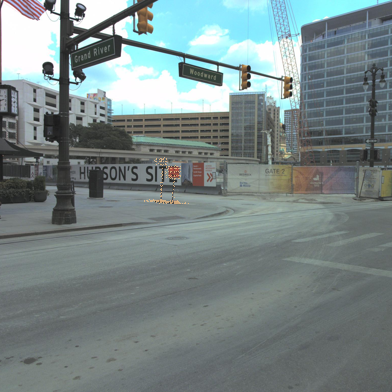}
    \caption{}
    \label{fig:supp-p_c}
  \end{subfigure}
  \hfill
  \begin{subfigure}{\gridwidth}
    \includegraphics[width=\imgwidth]{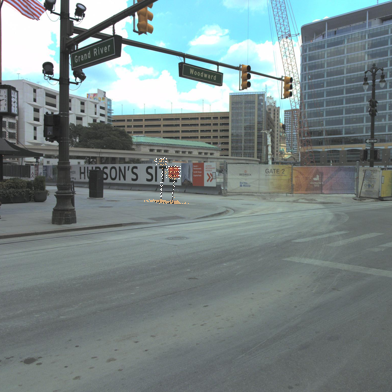}
    \caption{}
    \label{fig:supp-p_d}
  \end{subfigure}
  \hfill
  \begin{subfigure}{\gridwidth}
    \includegraphics[width=\imgwidth]{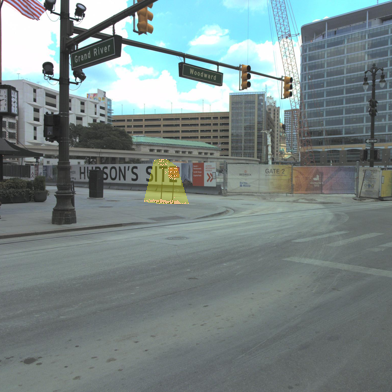}
    \caption{}
    \label{fig:supp-p_e}
  \end{subfigure}
  \\
  \begin{subfigure}{\gridwidth}
    \includegraphics[width=\imgwidth]{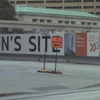}
    \caption{}
    \label{fig:supp-p_f}
  \end{subfigure}
  \hfill
  \begin{subfigure}{\gridwidth}
    \includegraphics[width=\imgwidth]{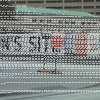}
    \caption{}
    \label{fig:supp-p_g}
  \end{subfigure}
  \hfill
  \begin{subfigure}{\gridwidth}
    \includegraphics[width=\imgwidth]{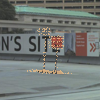}
    \caption{}
    \label{fig:supp-p_h}
  \end{subfigure}
  \hfill
  \begin{subfigure}{\gridwidth}
    \includegraphics[width=\imgwidth]{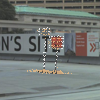}
    \caption{}
    \label{fig:supp-p_i}
  \end{subfigure}
  \hfill
  \begin{subfigure}{\gridwidth}
    \includegraphics[width=\imgwidth]{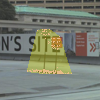}
    \caption{}
    \label{fig:supp-p_j}
  \end{subfigure}

  \caption{Step-by-step illustration of inferring image-space change maps from outdated 3D scenes. 
    (a) Raw RGB image. 
    (b) RGB image with synchronized LiDAR scan (white). 
    (c) RGB image with sparse $P_{\text{out}}$ change map (tangerline). 
    (d) Sparse change map (tangeline) after removing occluded or spurious detections (white) using LiDAR scans. 
    (e) Final convex hull of the filtered change points (yellow). 
    (f--j) Zoomed-in views of the corresponding steps in the first row.}
  \label{fig:supp-changemap}
\end{figure*}
\def\gridwidth{0.49\linewidth}

\begin{figure*}[ht]
    \centering
    \begin{tabular}{c c}

        \includegraphics[width=\gridwidth]{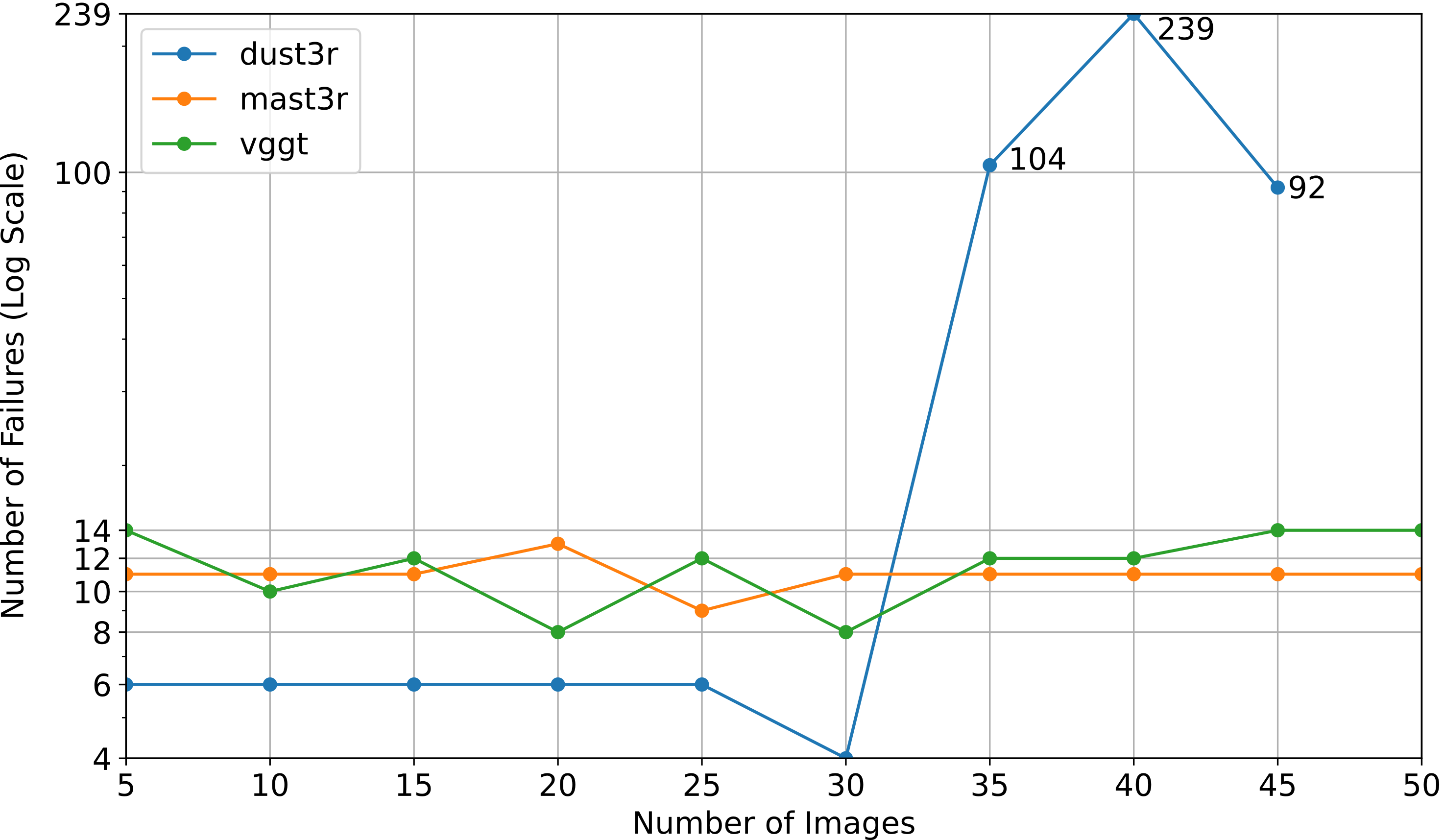} &
        \includegraphics[width=\gridwidth]{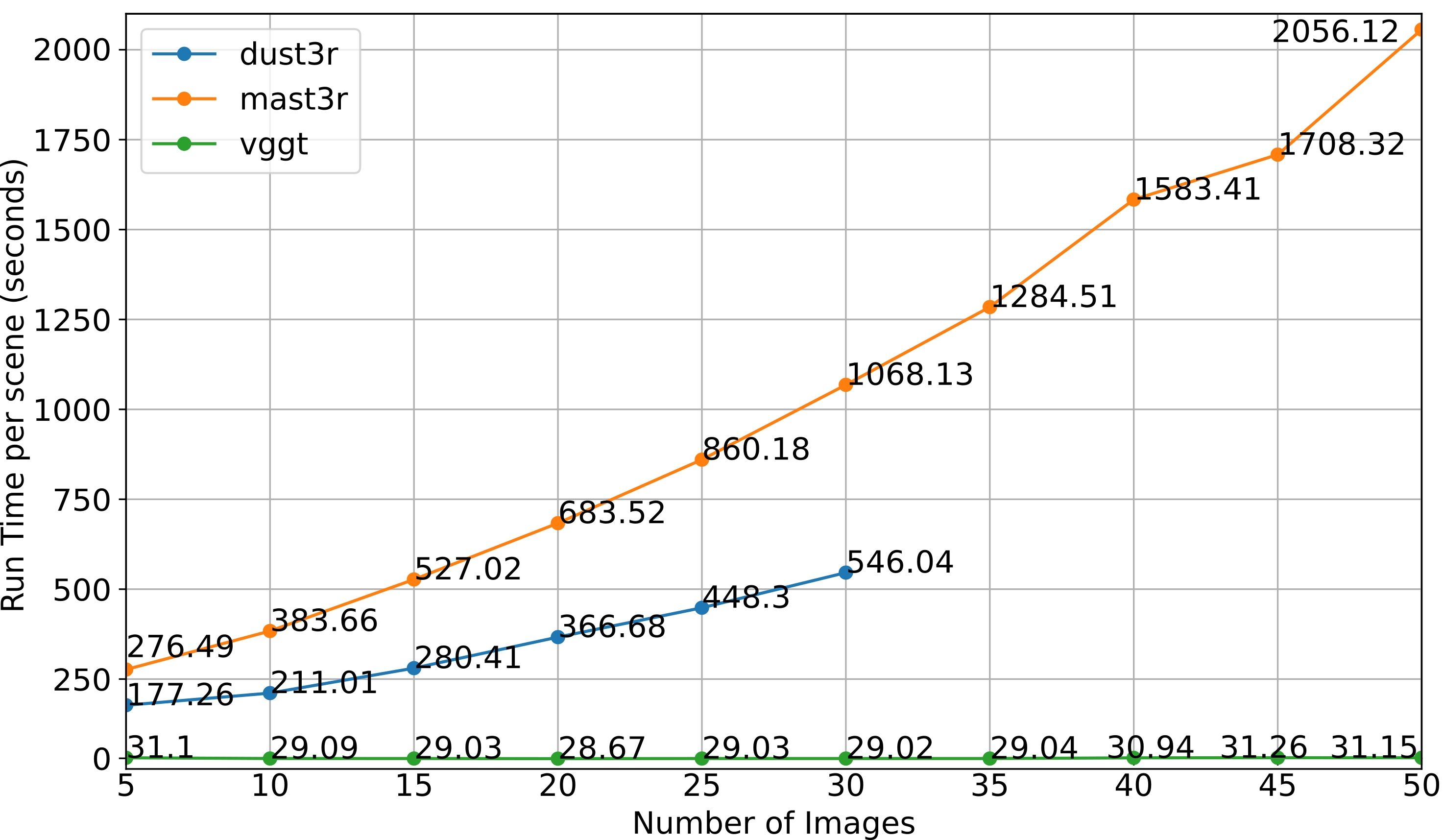} \\
        \includegraphics[width=\gridwidth]{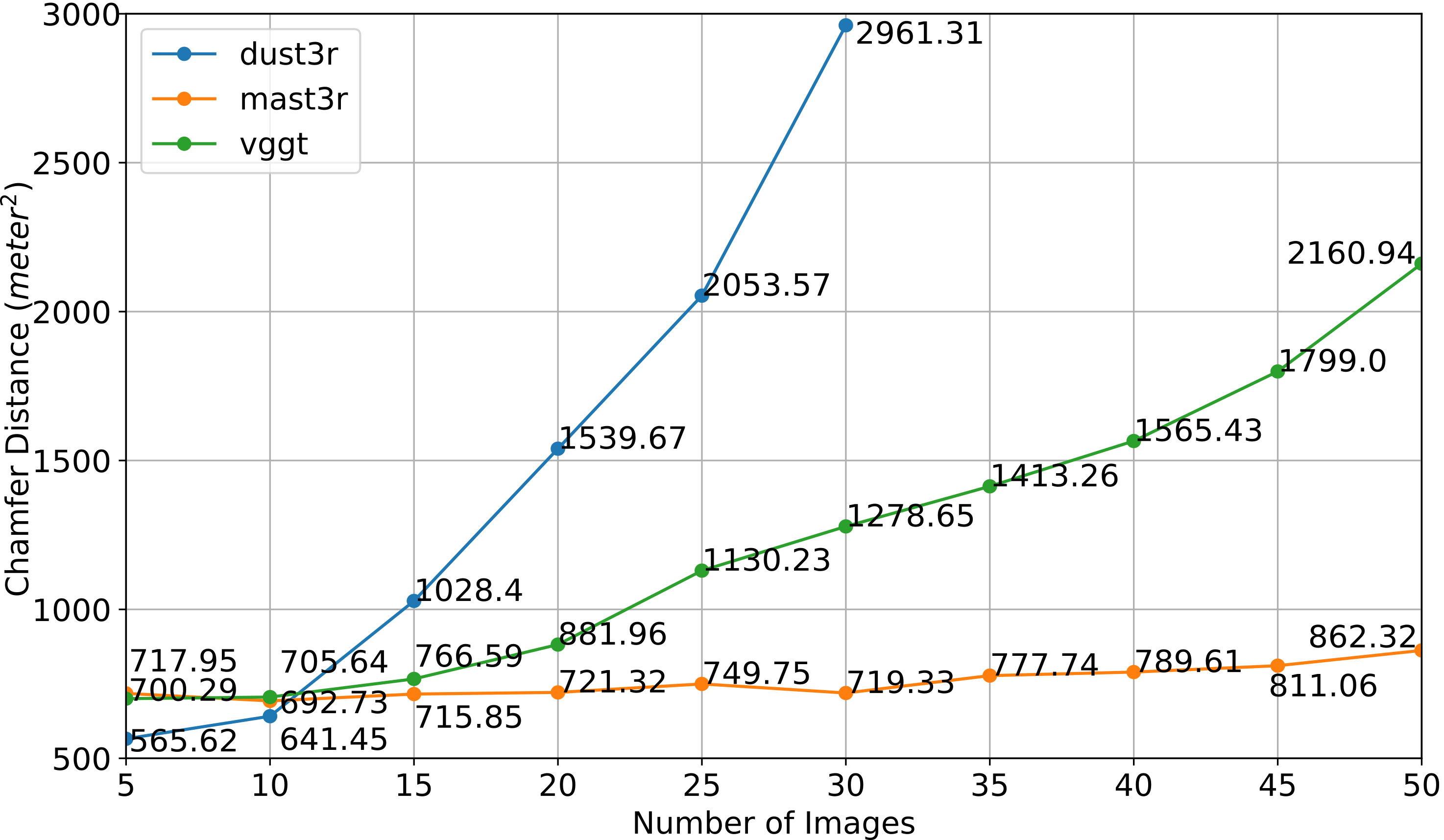} &
        \includegraphics[width=\gridwidth]{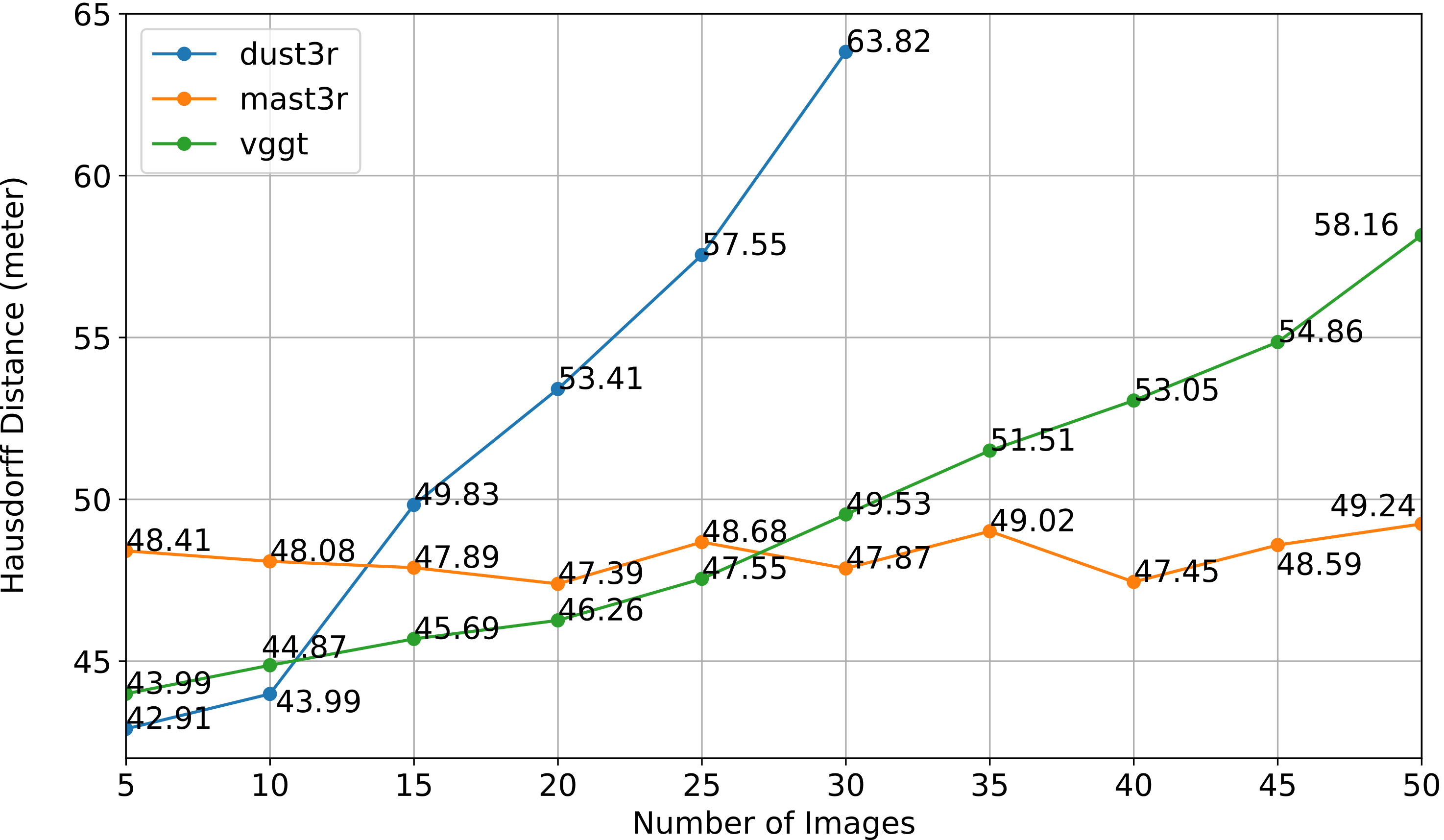} \\
        \includegraphics[width=\gridwidth]{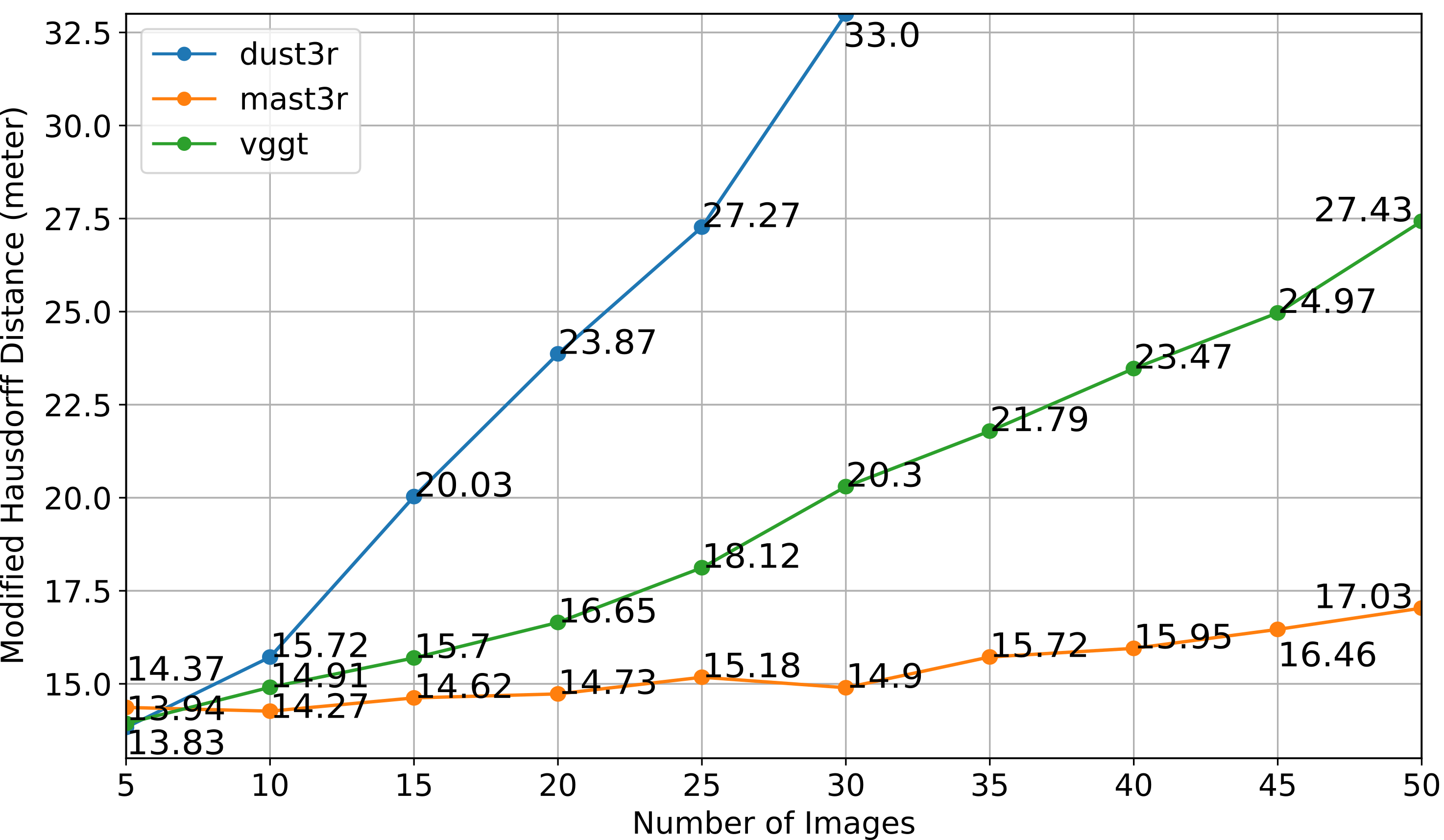} &
        \includegraphics[width=\gridwidth]{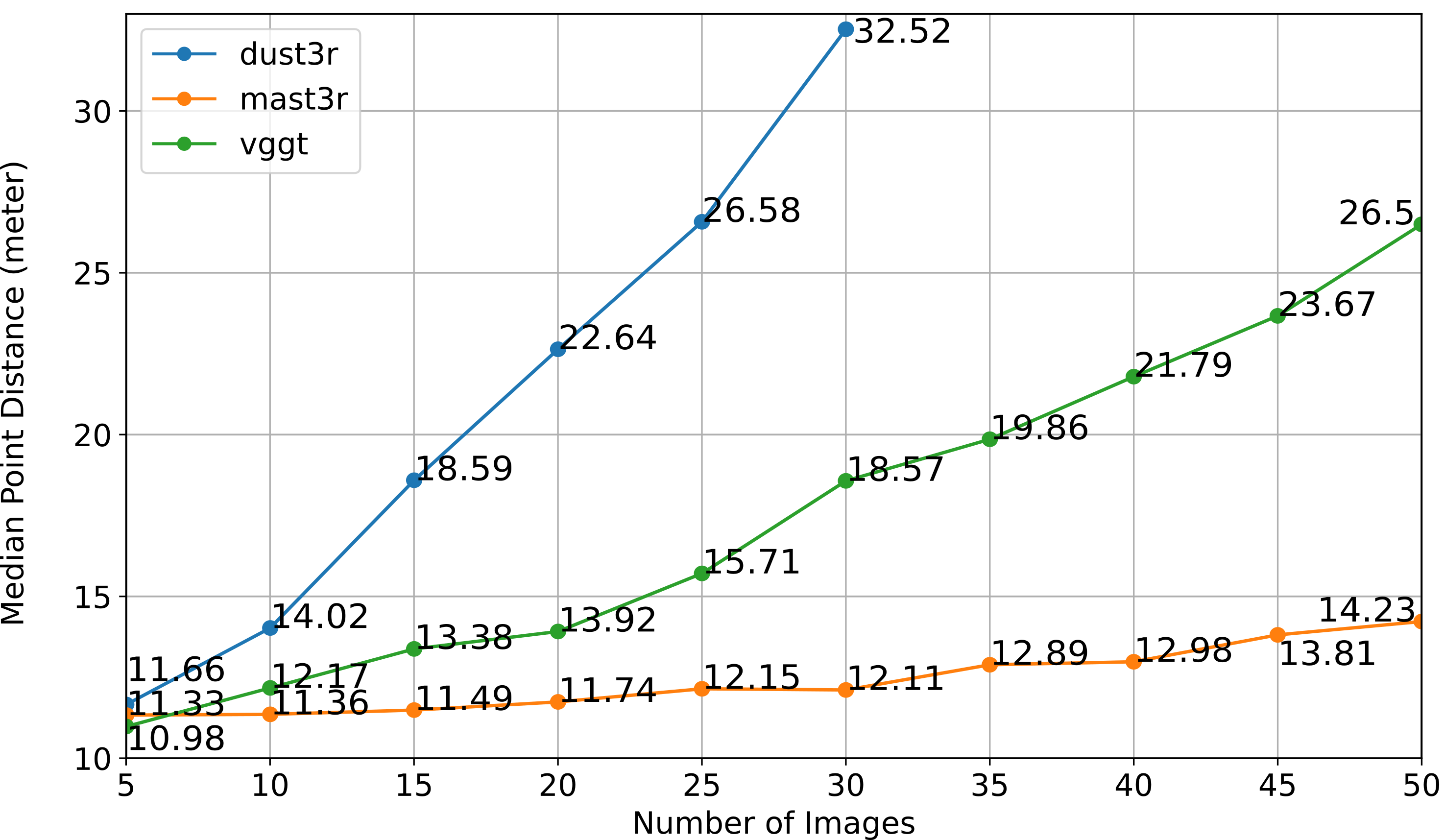} \\

    \end{tabular}
    \caption{
        Quantitative comparison of reconstruction performance across different numbers of input images for DUSt3R\cite{wang2024dust3r}, MASt3R\cite{leroy2024grounding}, and VGGT\cite{wang2025vggt}
        Metrics include: (top-left) number of reconstruction failures (log scale), (top-right) runtime per scene, (middle-left) Chamfer distance, (middle-right) Hausdorff distance, (bottom-left) modified Hausdorff distance, and (bottom-right) median point distance. Note that we do not report DUSt3R results beyond 35 input images due to excessive failure rates, as highlighted in the top-left plot.
    }
    \label{fig:supplement-addition}
\end{figure*}
\section{Supplemental: Static and Dynamic Labels}
\label{sec:supple-labels}

The label definitions used in this work are directly inherited from the Argoverse dataset \cite{chang2019argoverse, wilson2023argoverse}, which provides a comprehensive taxonomy of object categories commonly encountered in urban driving scenarios. We further reorganize these categories into two groups: \textit{dynamic} and \textit{static} to create \ours. The full lists of static and dynamic labels are listed in the following paragraph:

\paragraph{Static Labels (9 total)}

\begin{itemize}
  \item BOLLARD
  \item CONSTRUCTION\_BARREL
  \item CONSTRUCTION\_CONE
  \item MESSAGE\_BOARD\_TRAILER
  \item MOBILE\_PEDESTRIAN\_CROSSING\_SIGN
  \item OFFICIAL\_SIGNALER
  \item SIGN
  \item STOP\_SIGN
  \item TRAFFIC\_LIGHT\_TRAILER
\end{itemize}

\paragraph{Dynamic Labels (21 total)}

\begin{itemize}
  \item ANIMAL
  \item ARTICULATED\_BUS
  \item BICYCLE
  \item BICYCLIST
  \item BOX\_TRUCK
  \item BUS
  \item DOG
  \item LARGE\_VEHICLE
  \item MOTORCYCLE
  \item MOTORCYCLIST
  \item PEDESTRIAN
  \item RAILED\_VEHICLE
  \item REGULAR\_VEHICLE
  \item SCHOOL\_BUS
  \item STROLLER
  \item TRUCK
  \item TRUCK\_CAB
  \item VEHICULAR\_TRAILER
  \item WHEELCHAIR
  \item WHEELED\_DEVICE
  \item WHEELED\_RIDER
\end{itemize}

\section{Supplemental: Changes Objects in \ours}

To better characterize the changes introduced in \ours{}, we report the distribution of edited objects across both deletions and additions from 2,235 out-of-date scenes. \cref{fig:supp-del} summarizes the objects that were removed from the original static scenes $P^*_{upd}$. These deletions include both static objects annotated from Argoverse \cite{chang2019argoverse, wilson2023argoverse} and our newly annotated static objects (e.g., buildings, trees, tunnels, overpasses). Notably, our newly introduced annotations aim to bring important and more challenging change objects into the \ours{} dataset. In particular, these manually added labels correspond to large-scale or occlusion-heavy structures that are difficult to recover from a single image. For instance, the geometry of a tunnel or overpass is rarely visible in its entirety from any single viewpoint, and buildings often extend beyond the field of view of an individual frame. As a result, recovering these categories requires aggregating evidence across multiple images during reconstruction, which makes them especially valuable for benchmarking the robustness of change-detection and map-updating pipelines.

\cref{fig:supp-add} illustrates the objects that were added into the outdated scenes. These additions are drawn from our curated patch database, which contains pre-segmented object patches that can be reinserted into novel locations. Each object patch in the database has been manually inspected and cleaned to remove artifacts, incomplete geometry, and noisy boundaries. This ensures that when new objects are inserted, they integrate more cleanly into the voxelized point clouds and better reflect realistic scene changes. The distribution is dominated by frequently encountered road furniture such as stop signs, signs, and bollards, while less common categories like vehicular trailers or construction-related objects appear in smaller quantities. Together, these additions highlight common urban changes while maintaining high-quality insertions thanks to the manual refinement of the patch database.
\section{Supplemental: Change Maps}

While the sparse change map directly reflects the projected 3D change points, it is often too discontinuous to serve as reliable supervision for image-level tasks. Sparse detections make it difficult to delineate object boundaries and may leave large gaps that reduce label quality. By refining the sparse map into a dense change mask, we obtain spatially coherent regions that more accurately capture the extent of each changed object, enabling robust training and evaluation of image-based models.

To obtain the dense map \cref{fig:supp-p_j}, we further refine the sparse projections \cref{fig:supp-p_h} by checking the consistency of each pixel with the raw LiDAR scan \cref{fig:supp-p_g}. Specifically, we compare the depth of a projected change pixel against nearby LiDAR measurements in its local neighborhood. If the change point is likely occluded by closer geometry, it is removed \cref{fig:supp-p_i}. This process reduces spurious detections and ensures that only visible change regions are retained before generating the final dense masks.
\section{Supplemental: Point Addition Comparison}

To examine the effect of input image count on reconstruction, we conducted additional experiments across a range of image numbers for each method. The results in \cref{fig:supplement-addition} show that increasing the number of images does not improve robustness; instead, it often exacerbates failure rates and degrades accuracy. This suggests that simply feeding more images into these predictors does not necessarily yield more reliable reconstructions.

Based on these observations, we set the number of input images to 5 in the main paper. This choice minimizes the risk of instability while still allowing each method to function within its reliable regime. Using a small but sufficient number of images provides a fairer ground for comparison, avoids the severe failure cases observed at larger scales, and ensures that the reported benchmarks are representative of stable model performance.

\end{document}